\documentclass[accepted]{neutral} 
\usepackage[american]{babel}
\usepackage{natbib} 
\usepackage{mathtools}
\usepackage{booktabs}
\usepackage{tikz}

\usepackage{algorithm}
\usepackage{algorithmic}
\usepackage{graphicx} % Required for inserting images
\usepackage{amsmath}

\usepackage{centernot}
\usepackage{amssymb}
\usepackage{footnote}
\usepackage{geometry}
\usepackage{multirow}
\usepackage{booktabs}
\usepackage{bm}
\usepackage{algorithm}
\usepackage{algorithmic}
\usepackage{graphicx} % Required for inserting images
\usepackage{amssymb,amsthm}
\usepackage{graphicx}
\usepackage{subcaption}
\usepackage{mwe}
\usepackage{wrapfig}
\usepackage{cancel}

\newtheorem{theorem}{Theorem}
\newtheorem{definition}{Definition}
\newtheorem{assumption}{Assumption}

\newcommand{\indep}{\rotatebox[origin=c]{90}{$\models$}}

\title{Representation Learning Preserving Ignorability and Covariate
Matching \\
for Treatment Effects}

\author[1]{Praharsh Nanavati\thanks{Majority of this work was done when the author was at Indian Institute of Science Education and Research, Bhopal.}}
\author[2]{Ranjitha Prasad}
\author[3]{Karthikeyan Shanmugam}

\affil[1]{%
    Indian Institute of Science, Bangalore
}
\affil[2]{%
    Indraprastha Institute of Information Technology, Delhi
}
\affil[3]{%
    Google Deepmind, Bangalore
}
  
\begin{document}
\onecolumn
\maketitle
\begin{abstract}
Estimating treatment effects from observational data is challenging due to two main reasons; (a) hidden confounding, and (b) covariate mismatch (control and treatment groups not having identical distributions). Long lines of works exist that address only either of these issues. To address the former, conventional techniques that require detailed knowledge in the form of causal graphs have been proposed. Whereas, for the latter, covariate matching and importance weighing methods have been used. Recently, there has been progress in combining testable independencies with partial side information for tackling hidden confounding. \textit{A common framework to address both, hidden confounding and selection bias, is missing}. We propose neural architectures that aim  to learn a representation of pre-treatment covariates that is a valid adjustment and also satisfies covariate matching constraints. We combine two different neural architectures -- one based on gradient matching across domains created by subsampling a suitable \textit{anchor variable} that assumes causal side information followed by the other, a covariate matching transformation. We prove that approximately invariant representations yield approximate valid adjustment sets which would enable an interval around the true causal effect. In contrast to usual sensitivity analysis, where an unknown nuisance parameter is varied, we have a testable approximation yielding a bound on the effect estimate. We also outperform various baselines with respect to ATE and PEHE errors on causal benchmarks that include IHDP, Jobs, Cattaneo and an image-based Crowd Management dataset. The source code is available at \url{https://github.com/niftynans/matched_CFR}.
\end{abstract}

\section{Introduction}\label{sec:intro}
Machine Learning (ML) applications in healthcare, finance, and other safety-critical systems have driven the design of accurate and trustworthy decision-making algorithms. In this context, estimating the causal impact of an intervention is crucial. Although the availability of vast amounts of data has led to the design of causal inference (CI) techniques based on  observational data, compensating for the treatment bias and confounding factors in observational data continues to be a fundamental challenge \cite{sturm2014simple, torralba2011unbiased, jabri2016revisiting}. 

In scenarios when there is hidden confounding and when the underlying causal graph is known, graphical criteria such as \textit{backdoor criterion} can be used to identify \textit{valid adjustment sets} addressing the issue of confounding bias. In the case of backdoor adjustment, regressing the outcome on the treatment and covariates in these sets predicts the interventional effect. This occurs without bias in the sample limit \cite{pearl2009causality}.  %Recent works have focused on the structure learning of causal graphs \cite{zheng2018dags, lachapelle2019gradient, yu2019dag}.
However, in real-world scenarios, it is often not the case that the causal graph, that also specifies the pairs of variables that are confounded, can be easily and accurately determined. 

On the other hand, when one assumes \textit{strong ignorability} (or informally no effective hidden confounding), several ML-based approaches for individual treatment effect estimation view the causal inference problem as a covariate shift problem between treatment and control group. They propose representation learning algorithms that balance the factual and the counterfactual population \cite{shalit2017estimating, louizos2017causal, curth2023using} in distribution. One of the main drawbacks of these works is the assumption of strong ignorability (Definition \ref{def: 3}), a conditional independence assumption involving unseen counterfactuals.

In the realm of domain generalization, it is essential that predictive models generalize to unseen domains. A key aspect of domain generalization is to suppress the features that are spuriously correlated with the labels and learn from features that are invariant across multiple domains \cite{lu2021nonlinear, magliacane2018domain, zhang2015multi, kamath2021does, arjovsky2019invariant, shi2021gradient}. This can be enabled by learning nonlinear, invariant, causal predictors from various environments. The notion of Invariant Risk Minimization (IRM) in \cite{arjovsky2019invariant} emphasizes the invariance view of causation, where an intermediate representation is learnt which helps in obtaining a subsequent domain-invariant predictor. In \cite{shi2021gradient}, the authors propose to impose invariance using an inter-domain gradient matching penalty. %However none of the above mentioned works address the problem of individual treatment effect estimation while learning the features that remain invariant across the domains.

In another line of work, \cite{shah2022finding,cheng2022toward,entner2013data} showed that when ignorability is not assumed for pre treatment variables, if one of the features is known to be a direct parent of the treatment, an \textit{anchor variable}, if an independence test involving this anchor variable, outcome and a subset of pre treatment variables $\mathbf{z}$ passes, then $\mathbf{z}$ is a valid backdoor. \cite{shah2022finding} showed that IRM oracles can be used with a suitable sub sampling procedure involving the anchor variable which artificially splits the observational data into multiple domains.

In this work, we address the following question: \textit{Under the anchor variable assumption, is it beneficial to learn a representation of pre-treatment variables not involving the anchor variable that is a valid adjustment and simultaneously satisfies covariate matching constraints?} \\ 
% \KS{ Contributions needs to be rewritten. But Introduction done till here.}

\subsection{Our Contributions} 

In this work, we address the problem of individual treatment effect estimation by obtaining invariant features across domains. Our contributions are as follows: 
\begin{itemize}
    \item We propose a novel neural architecture that learns representation which accurately estimates treatment effects in the presence of hidden confounding as well as satisfying covariate matching constraint across treatment groups. The learnt representation reduces the effects of confounding as it incorporates a second-order linear approximation of inter-domain gradient matching (IDGM) known as  the FISH algorithm \cite{shi2021gradient}. Sequentially, we refine the representation by incorporating covariate matching constraints using an Integral Probability Metric (IPM) (e.g., maximum mean discrepancy, Wasserstein distance, etc) to ensure a balanced representation \cite{shalit2017estimating}. Since we deal with real-world datasets where domain knowledge often plays a huge role, some partial structural knowledge (such as the anchor variable assumption) is required for the general, non-ignorable case. For instance, in the case of healthcare, only one feature that is used in the treatment protocol (such as age or some specific health condition like diabetes etc.) needs to be known.
    \item We generate multiple domains within the data for our work using an \textit{anchor variable}, that is known to be an observed parent of the treatment variable \ref{algo_dom_generation}, similar to  \cite{shah2022finding} with invariance testing. We demonstrate that the proposed technique works universally across multiple datasets and modalities. Apart from the benchmark datasets, we have also shown experimental results on the crowd management dataset\cite{takeuchi2021grab}. This work focuses on the problem of estimating the cost of crowd movement (such as evacuation times etc) from a relatively small amount of biased data. The data is modeled based on a simulated situation of crowds, such as their current locations, which are observed as covariates.
\end{itemize}

\subsection{Why is this non-trivial?} 
    Both oracles (Inter Domain Gradient Matching and Covariate matching) serve different purposes. Covariate Matching needs to assume full ignorability (TARNet, CFRNet \cite{johansson2016learning, johansson2018learning}). Covariate matching is done to remove treatment bias, which is due to the unknown policy. Having IDGM (Inter Domain Gradient Matching, an IRM proxy) as an extra module to tackle the effects of hidden confounding is essential as it helps in obtaining approximate valid adjustment sets while assuming non-ignorability. Combining these oracles is also not straigthforward as we require partial side-information of the underlying causal graph in order to enable the anchor variable formulation where, IRM-based objectives applied on the rest of the covariates without the anchor achieves ignorability if there is perfect invariance. This is shown by \cite{shah2022finding}. We rely on this result to demonstrate that sequential application of the two representation learning algorithms would result in balanced and ignorable representation of treatment effect estimation. However \cite{shah2022finding} use \cite{arjovsky2019invariant}'s regularizer which may not permit general representation learning (the invariant predictor on top has to be linear). Hence, we adopt a well known IRM proxy –- IDGM (shown to work well in OOD benchmarks like \cite{koh2021wilds}) that helps in reducing the effects of hidden confounding using the anchor variable approach, but permits very general invariant predictors and representations. This weakens the principal untestable assumption of ignorability of the given covariates made in the CFRNet line of work.  We show later in our experiments, that the sequential application of these oracles outperforms their alternating combination. We also show that approximate invariance is sufficient to get an interval around the true treatment effect.
    \begin{quote}
        \textit{To the best of our knowledge, our work is the first to consider this stronger goal of isolating the causal effect of the treatment effect on the outcome while assuming non-ignorability of the provided pre-treatment covariates.}
    \end{quote}

We discuss related works in Section \ref{sec:related} followed by the problem formulation in Section \ref{sec:problem} where we showcase our working algorithm and experiments and other results in Section \ref{sec:experiments}.

\begin{figure*}[t]
\centering
  \includegraphics[width=16cm]{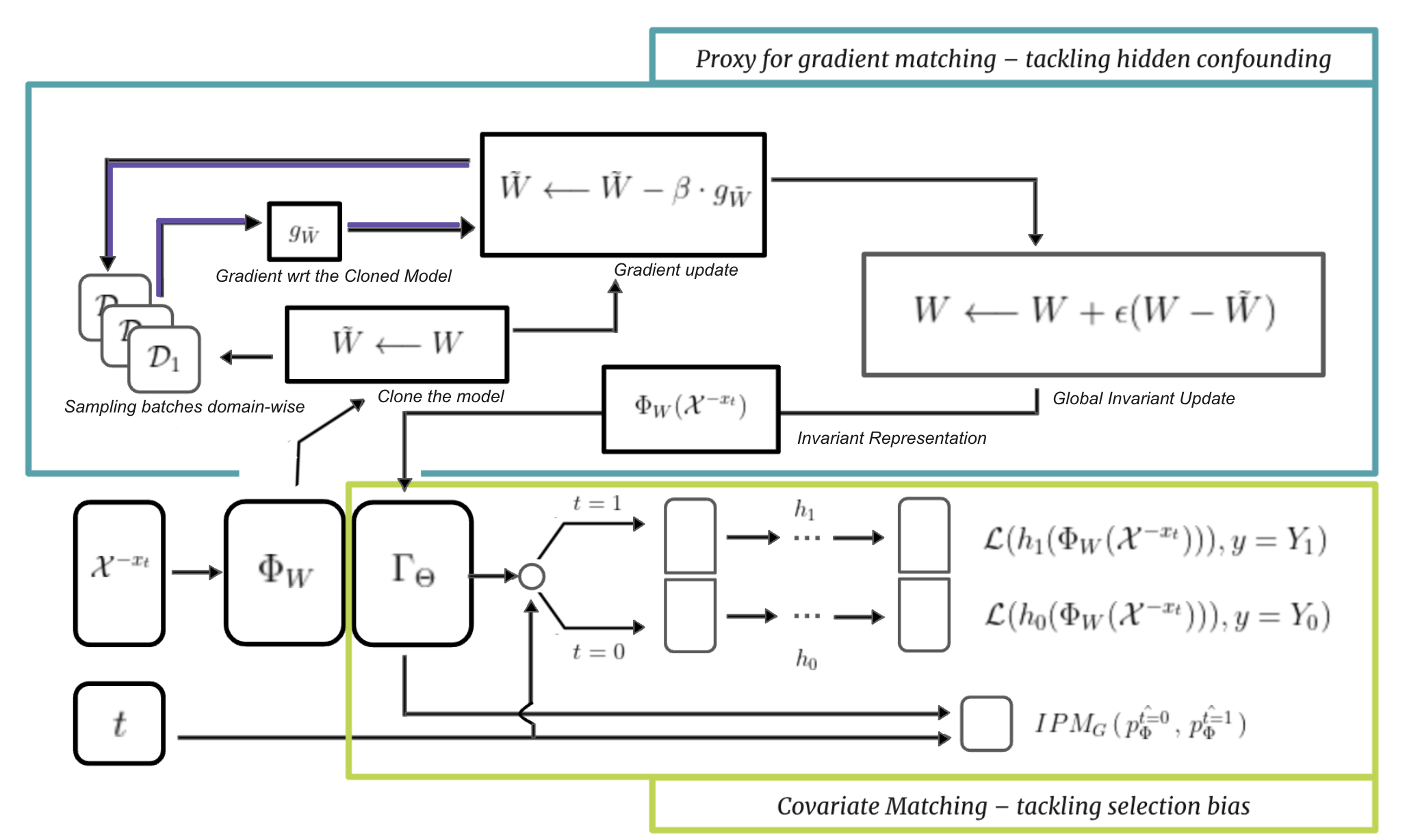}
  \caption{Our proposed architecture. Here, $\mathcal{X}^{-X_t}$ denotes covariates without the anchor variable, i.e., an instance of the features recorded in the dataset, $t$ is the treatment variable, and $Y_0$ and $Y_1$ are the factual outcomes according to the group to which the instance belongs to.
  $W$ are the weights of the initial Matching Representation network, which are updated by the FISH Algorithm \cite{shi2021gradient} -- a second order linear approximation of IDGM. The data is divided into domains and then the gradient inner product is used to update $\tilde{W}$. Once the final weights, $\tilde{W}$ are learnt across domains and across batches, we pass on $\Phi_{W}(\mathcal{X}^{-X_t})$ as covariates in a new projection space to the CFR model, $\Gamma_{\Theta}$ \cite{shalit2017estimating}. The hypothesis layers $h_1$ and $h_2$ learn the treated and the control distributions separately, while the $IPM$ is applied to the treated and the control distributions in the projection space. Both objectives, invariance \cite{shi2021gradient}, and covariate matching \cite{shalit2017estimating} are achieved before estimating the causal effect.}
  \label{fig:ourarch}
\end{figure*}

\section{Related Works} \label{sec:related}
Individual treatment effect estimation aims to predict the treatment outcomes of an individual under different interventions, and  counterfactual inference is an important tool in this direction. Traditional propensity score matching or importance weighting \cite{hirano2003efficient,rosenbaum1983central} are limited in their ability in the presence of confounding variables and high dimensional covariates. Most of these works also assume strong ignorability (Definition \ref{def: 3}). Learning balanced representations is a flexible technique to mitigate the impact of treatment bias \cite{johansson2016learning,yoon2018ganite} while being efficient in high dimensions. In particular, \cite{shalit2017estimating} derives the generalization-error bound for
estimating individual causal effect while proposing the CFRNet architecture to optimize the bound. Despite the empirical success of such methods, it has been emphasized that over-enforcing balance can be harmful \cite{alaa2018limits}. Alternate approaches have also tackled a slightly different problem of model generalization using balanced representations under design shift \cite{johansson2018learning}.

Traditional approaches assume complete availability and knowledge of the underlying causal graph (DAG) and provide ways to determine valid adjustments using graphical criteria such as the backdoor and frontdoor criteria using d-separation \cite{tian2002general, shpitser2008complete} properties of the causal graph. Recent works have tried working under assumptions of partial knowledge of the underlying causal graph \cite{entner2013data, cheng2022toward, shah2022finding}.

Recently, Domain generalization methods address the issue of ML model performance on unseen, out-of-distribution data in spite of being trained on a limited number of seen domains. In particular, Invariant Risk Minimization (IRM) focuses on learning representations of data that yield an invariant optimal predictor on top of it across domains \cite{arjovsky2019invariant}. Various oracles for performing IRM has been proposed \cite{ahuja2020invariant,shi2021gradient,arjovsky2019invariant}. We use \cite{shi2021gradient} as a component, suitably modified in our work to impose invariance of a certain kind to ensure ignorability of our representations.

\section{Problem Formulation}\label{sec:problem}
\subsection{Problem Setup and Assumptions}

Consider the observational training data comprising $\{\mathcal{X}, T, Y\}$  where $\mathcal{X}$ represents the set of all covariates, $T$ is the binary treatment variable and $Y$ is the outcome. 

We begin by defining a Semi-Markovian Causal Model and what an intervention is.

\begin{definition}
\cite{tian2002general,shah2022finding}  A Semi-Markovian Causal Model is a tuple $<\mathcal{V}, \mathcal{U}, \mathcal{G}, \mathbb{P}(v | \mathrm{Pa}^{o}(v), \mathrm{Pa}^{u}(v)),\mathbb{P}(\mathcal{U}) >$ such that a) $\mathcal{V}$ is the set of observed variables, b) $\mathcal{U}$ is the set of unobserved variables
c) $\mathcal{G}$ is the DAG over the set of vertices $\mathcal{V} \cup \mathcal{U}$ such that each member in $\mathcal{U}$ has no parents and at-most two children. d) $\mathbb{P}(v | \mathrm{Pa}^{o}(v), \mathrm{Pa}^{u}(v)) \forall v \in \mathcal{V}$ is the set of conditional distributions of the observed variables given observed and unobserved parents. $\mathrm{Pa}^{o}(v) \subset \mathcal{V} $ and $\mathrm{Pa}^{u}(v) \subset \mathcal{U} $ are the observed and unobserved parents and e) $\mathbb{P}(\mathcal{U})$ is the joint distribution over the unobserved features.
\end{definition}
\begin{definition}
\cite{brouillard2020differentiable, pearl2009causality} A \textit{do} intervention on a variable $v$ setting it to value $x$ (denoted $do(v=x)$) corresponds to replacing its conditional $\mathbb{P}(v | \mathrm{Pa}^{o}(v), \mathrm{Pa}^{u}(v))$ by $\mathbf{1}_{v=x}$, modifying the distribution only locally by an indicator function $\mathbf{1}_{(\cdot)}$. This distribution is called `do-distribution' or the intervenional distribution.
\end{definition}

\begin{assumption}
We assume that we have a Semi-Markovian Causal Model over $\langle T, Y, \mathcal{U}, \mathcal{X} \rangle$ where $\mathcal{X}$ is the set of pre-treatment variables, $T$ is the treatment, $T \rightarrow \mathcal{Y}$ exists, and $Y$ has no children and ${\cal U}$ is unobserved.
\label{ass:a1}
\end{assumption}

We denote, $Y(1) \sim  \mathbb{P}_{T \leftarrow 1}(Y)$ as the outcomes of samples which belong to the treatment group, i.e., where $T = 1$ and $Y(0) \sim  \mathbb{P}_{T \leftarrow 0} (Y)$ as the outcomes of samples in the control group, or $T = 0$. We now define ignorability, which effectively assumes away hidden confounders.

\begin{definition} \label{def: 3}
    \cite{rubin2005causal, shalit2017estimating, johansson2016learning} The Strong ignorability assumption states that,
    $\, Y(0) , Y(1) \, \indep \, T \, | \, \mathcal{X}$
\end{definition}
 
To estimate our causal effect accurately, we wish to condition not on the entire set of covariates but only on a subset of features that lie in the valid adjustment set. One graphical criterion popular in the literature is the backdoor criterion given below.

\begin{quote}
    \textit{Given an ordered pair of variables $(A, B)$ in a directed acyclic graph $\mathcal{G}$, a set of variables $\mathcal{Z}$ satisfies the backdoor
    criterion relative to $(A,B)$ if no node in $\mathcal{Z}$ is a descendant of $A$, and $\mathcal{Z}$ blocks every path between $A$ and $B$ that contains an arrow into $A$} \cite{tian2002general, pearl2009causality, shah2022finding}.
\end{quote}

\textbf{Remark:} The works blocking above refers to the notion of d-separation which is a graphical separation criterion in causal graphs that imply conditional independencies (we refer the reader to the supplement for such a definition).

\begin{algorithm}[t!]
    \caption{Counterfactual Regression with Covariate Matching and Valid Adjustment (${\cal S}, {\cal D},\alpha,\beta,\lambda,\epsilon$)}\label{algo_dom}
    \begin{algorithmic}
        \STATE \textbf{Input:} Factual samples, ${\cal S} = (x_1, t_1, y_1), ... , (x_n, t_n, y_n)$. Set of domains, $\mathcal{D} = \{\mathcal{D}_1, \mathcal{D}_2,..., \mathcal{D}_s\}$  - a partition of ${\cal S}$ using sub-sampling variable $e$ from \ref{algo_dom_generation} on ${\cal S}$, $\alpha$ - IPM (Integral Probability Metric) loss regularizer, $\eta$ - learning rate, $\mathcal{L}(\cdot, \cdot)$ - Mean Squared Loss error. $\beta$, $\lambda$ and $\epsilon$ are hyperparameters. $\mathcal{X}^{-X_t}$ denotes the covariate $x$ without the anchor variable. 
        \vspace{2pt}
        \STATE \textbf{Initialize:} Gradient Matching Representation network $\Phi_{W}$ ( with initial weights $W$), CFR Representation network $\Gamma_{\Theta}$ (with initial weights, $\Theta$) outcome network $h_V$ (with initial weights, V). 
        \STATE Compute $u$ = $(\sum 
        \limits_{i = 1}^{n} t_{i}) / n $, $ w_i$ = $\frac{t_i}{2u} + \frac{1-t_{i}}{2 (1- u)}$ for j = 1,.., $n$.
        \STATE \texttt{Gradient Matching:}\FOR{iterations = $1, 2,...$ to $N$}
            \STATE $\tilde{W} \longleftarrow W$
            \FOR{$\mathcal{D}_k \in$ permute $(\{\mathcal{D}_1, \mathcal{D}_2,..., \mathcal{D}_s\})$}
                \STATE Sample batch $d_k \sim \mathcal{D}_k$, Let $m$ = $|d_k|$
                \STATE %$g_{\tilde{W}} = \nabla_{\tilde{W}} 
                %\frac{1}{m} \sum \limits_{j \in d_k} w_{i,j} \mathcal{L}(h_v(\Phi_{\tilde{W}}(x_{ij}), t_{ij}), y_{ij})$
                $g_{\tilde{W}} = \nabla_{\tilde{W}} 
                \frac{1}{m} \sum \limits_{j \in d_k} w_{j} \mathcal{L}(h_V(\Phi_{\tilde{W}}(x^{-X_t}_{j}), t_{j}), y_{j})$
                %\STATE$\tilde{G_{k}} = \mathbf{E}_{d_k}[g_{\tilde{W}}] $ \KS{What is this step ? }
                \STATE $ \tilde{W} \longleftarrow \tilde{W} - \beta  g_{\tilde{W}}$
            \ENDFOR \\

            $W \longleftarrow W + [\epsilon(W - \tilde{W})]$
        \ENDFOR
        \STATE \texttt{Covariate Matching:}\WHILE{not converged}
        \STATE Sample mini-batch $\{i_1, i_2, .., i_m\} \in \{\ 1,2,..n\}$
        \STATE Calculate the empirical losses and the IPM loss;
        \STATE \small$g_{V} = \nabla_{V} \frac{1}{n} \sum \limits_{j=1}^m w_{i_j} \mathcal{L}(h_V(\Gamma_{\Theta}(\Phi_{W}(x^{-X_t}_{i_j})), t_{i_j}), y_{i_j})$
        \vspace{2pt}
        \STATE \small$g_{\Theta} = \nabla_{\Theta} \frac{1}{n} \sum \limits_{j=1}^m w_{i_j} \mathcal{L}(h_v(\Gamma_{\Theta}(\Phi_{W}(x^{-X_t}_{i_j})), t_{i_j}), y_{i_j})$
        \vspace{2pt}
        \STATE $\mathcal{G}_0 \leftarrow \{\Gamma_{\Theta}(\Phi_{W}(x^{-X_t}_{i_j}))\}_{t_{i_j} = 0}$
        \vspace{2pt}
        \STATE $\mathcal{G}_1 \leftarrow \{\Gamma_{\Theta}(\Phi_{W}(x^{-X_t}_{i_j}))\}_{t_{i_j} = 1}$ 
        \vspace{2pt}
        \STATE \small$g_{IPM} = \nabla_{\Theta} IPM_{G}(\mathcal{G}_0, \mathcal{G}_1)$.
        \vspace{6pt}
        \STATE \small$[\Theta, V] \longleftarrow [\Theta - \eta(\alpha*g_{IPM} + g_{\Theta}), V - \eta(g_{V} + 2\lambda V)]$ 
        \ENDWHILE
        \STATE \textbf{Output:} [$\Gamma_\Theta, \phi_W, h_V$]
    \end{algorithmic}
\end{algorithm}

Now, if any $\mathcal{Z} \subseteq \mathcal{X}$ such that $\,(\,Y_0\,,\, Y_1\,)\, \indep \, T \,| \,\mathcal{Z} \,$, then $\mathcal{Z}$ satisfies the ignorability condition. When $\mathcal{Z}$ is ignorable, it is also a \textit{valid adjustment} set with respect to $(T, Y)$ and hence the ATE can be estimated by regressing on $\mathcal{Z}$ \cite{shah2022finding} using the adjustment formula $\mathbb{E}[Y|do(T=t)] = \sum \limits_{z} \mathbb{E}[Y | T=t, {\cal Z} =z]$. If ${\cal Z}$ is backdoor then it is also a valid adjustment set and satisfies ignorability condition.

\begin{theorem} \label{thm: 4}
    \cite{shah2022finding} Let Assumption \ref{ass:a1} be satisfied. Consider any $X_t \in \mathcal{X}$, that has a direct edge to $T$ called the anchor variable. Let $E$ be sub-sampled using $X_t$ , i.e., $e = f(X_t, \eta)$  where $\eta$ is a noise variable independent of the number of vertices in the graph, and $f$ is a function of $X_t$ (realization of $X_t$) and $\eta$. Let $\mathcal{X}^{-X_t}$ denote all features without $X_t$ and its realization is denoted $x^{-X_t}$. Then, under faithfulness assumptions, if for any ${\cal Z} \subseteq {\cal X}^{-X_t}$ $Y \indep E | (T, {\cal Z})$, then ${\cal Z}$ is a valid backdoor and a valid adjustment set.
\label{thrm:theorem_1}
\end{theorem}

\begin{theorem} \textbf{(Our Result)}
Let Assumption \ref{ass:a1} be satisfied. Considering all vairable definitions from Theorem \ref{thm: 4}, and under the faithfulness assumption, we can say that $\epsilon$-approximate invariance implies the presence of approximate valid adjustments in the case of a Linear SEM, i.e. for some $\epsilon > 0$,
$$cov(Y, X_t |T, Z) < \epsilon \,\, \implies cov(Y', T | Z) < \epsilon$$
where $Y'$ denotes the post-interventional distribution of $Y$.
Proof is given in Appendix \ref{sec:proof}.
\label{thrm:our_theorem}
\end{theorem}
In other words, invariance of the outcome conditioned on subsets of observed features (not containing the known anchor) and the treatment across environments created using $X_{t}$ is equivalent to checking for valid backdoors. Hence, to impose conditions stated in Theorem \ref{thrm:theorem_1}, we can use an Invariant Risk Minimization Oracle \cite{arjovsky2019invariant} as was suggested in \cite{shah2022finding}. Empirically one finds $\phi({\cal X}^{-X_t})$ such that $Y \indep E| T, \phi({\cal X}^{-X_t}) $ where $\phi$ is an invariant representation learnt using an IRM oracle. However, in order to achieve this, the presence of multiple domains ($E$) in the dataset is necessary. Hence, we define domains within the data using the sub-sampled variable $E$ defined in Theorem \ref{thrm:theorem_1} using Algorithm \ref{algo_dom_generation}. However, far from just learning IRM representations we focus on the following twin objectives. 
\\

\subsection{Combining IDGM and Covariate Matching}
\label{Rep_CFR}
We are interested in estimating the causal effects of a treatment on a specific outcome in the presence of heavy confounding. We aim to build a representation for causal effect estimation that aims to use invariance to ensure valid adjustment in the presence of confounding while using a domain adaptation type regularizer to avoid mismatch in covariate distribution across treatment groups. 
% \begin{quote}
%     \textit{Can we obtain a representation learner that can give us the best of both these worlds?}
% \end{quote}
Using the subsampled environmental variable $E$, we generate domains, $\mathcal{D} := \{\mathcal{D}_1, \mathcal{D}_2, ..., \mathcal{D}_S \}$ 
 for our factual dataset,  $\{\mathcal{X}, T, Y\}$. Each factual sample in the dataset is mapped to a domain $\mathcal{D}_i \in \mathcal{D}$. Since we use the anchor variable $X_t$ to do so, we obtain our set of domains $\mathcal{D}$, and our factual samples can now be denoted as $\{\mathcal{X}^{-X_t}, T, Y\}$.
 
\noindent \textbf{IDGM}: IDGM is an approximate IRM  oracle desgined to learn an invariant predictor. To perform IDGM, if we are given a model $W$ and loss function $\ell$, the expected gradients for data across the $n^{th}$ domain, where $n \leq s$ would be:
\begin{align}
\mathcal{G}_n = \mathbb{E}_{\mathcal{D}_n}\frac{\partial l((x, y); W)}{\partial W}
\label{gn}
\end{align}
If gradients of two domains $i$ and $j$ point in a similar direction, i.e., if $\mathcal{G}_i \cdot \mathcal{G}_j > 0$, taking a gradient step along either of these gradients improves the model's performance on both these domains. This can help in learning invariant features, and spurious correlations are subsequently removed. This is what is called as Inter Domain Gradient Matching (IDGM).
Since direct optimization of the gradient inner product can be computationally prohibitive, it requires computation of second-order derivatives  \cite{shi2021gradient}. 
As we show in \ref{Rep_CFR}, this procedure can be used to produce a matched representation function which is better suited for covariate matching for causal effect isolation. The IDGM loss which performs IRM, is defined as,
\begin{align}
\mathcal{L}_{idgm} = \mathcal{L}_{erm}(\cup {\cal D}_i; W) - \gamma\frac{2}{S(S-1)}\Sigma^{i \neq j}_{i,j \in S}(\mathcal{G}_i \cdot \mathcal{G}_j).
\end{align}

\noindent \textbf{Covariate Matching:}~To solve the issue of treatment bias, representations that match treatment and control distributions is often used, i.e., if $P_{t}(\Gamma_\Theta(\Phi(\mathcal{X}))) \sim P_{c}(\Gamma_\Theta(\Phi(\mathcal{X})))$ when $\mathcal{X}$ is ignorable with a one-to-one function $\Phi(\cdot)$. Covariate matching can be achieved by enforcing an IPM, such as Maximum Mean Discrepancy or Wasserstein Distance \cite{shalit2017estimating} between the two distributions.

\subsection{Our Algorithm}
We perform Inter-Domain Gradient Matching and Covariate matching sequentially on ${\cal X}^{-X_t}$ to make the resulting representation covariate matches and satisfy ignorability assuming $X_t$ satsifies the anchor variable assumption in \ref{thrm:theorem_1}. Once we learn our representations $\Phi_{W}$ using IDGM, we pass the transformed covariates to the CFR architecture \cite{shalit2017estimating} to learn  $\Gamma_{\Theta}$ along with $t$ to perform covariate matching using IPM scaling as done in \cite{shalit2017estimating, johansson2022generalization} using a Neural Net architecture. We define our architecture in Figure \ref{fig:ourarch} and describe the process in detail in Algorithm \ref{algo_dom}. We show through experiments that this combined architecture is very effective empirically.
Theorem \ref{thrm:our_theorem} shows that approximately invariant representations yield approximate valid adjustment sets, enabling computing an interval around the true causal effect isolation. In contrast to usual sensitivity analysis, where an unknown nuisance parameter is varied, we have a testable approximation yielding a bound on the effect estimate. IDGM in pratice ensures approximately invariant representations ($\epsilon$-approximate invariance) in the first step of our algorithm is sufficient to provide an interval estimate on the true effect. 
% \begin{figure*}[h!]
%     \centering
%     \includegraphics[width = \linewidth]{figures/IHDP_results.png}
%     \caption{Since we are interested in the generalisability of our method, we showcase the ATE and PEHE errors for out-of-distribution samples for the IHDP dataset in the plot above. There is a significant overlap in the ATE errors, the PEHE errors (for the harder problem of individual treatment effect estimation) for our method are considerably lower than the baselines. This establishes that obtaining an ignorable representation is vital over and above covariate matching for individual treatment effects estimation.}
%     \label{fig:IHDP}
% \end{figure*}

\begin{figure*}[h!]
    \centering
    \includegraphics[width = \linewidth]{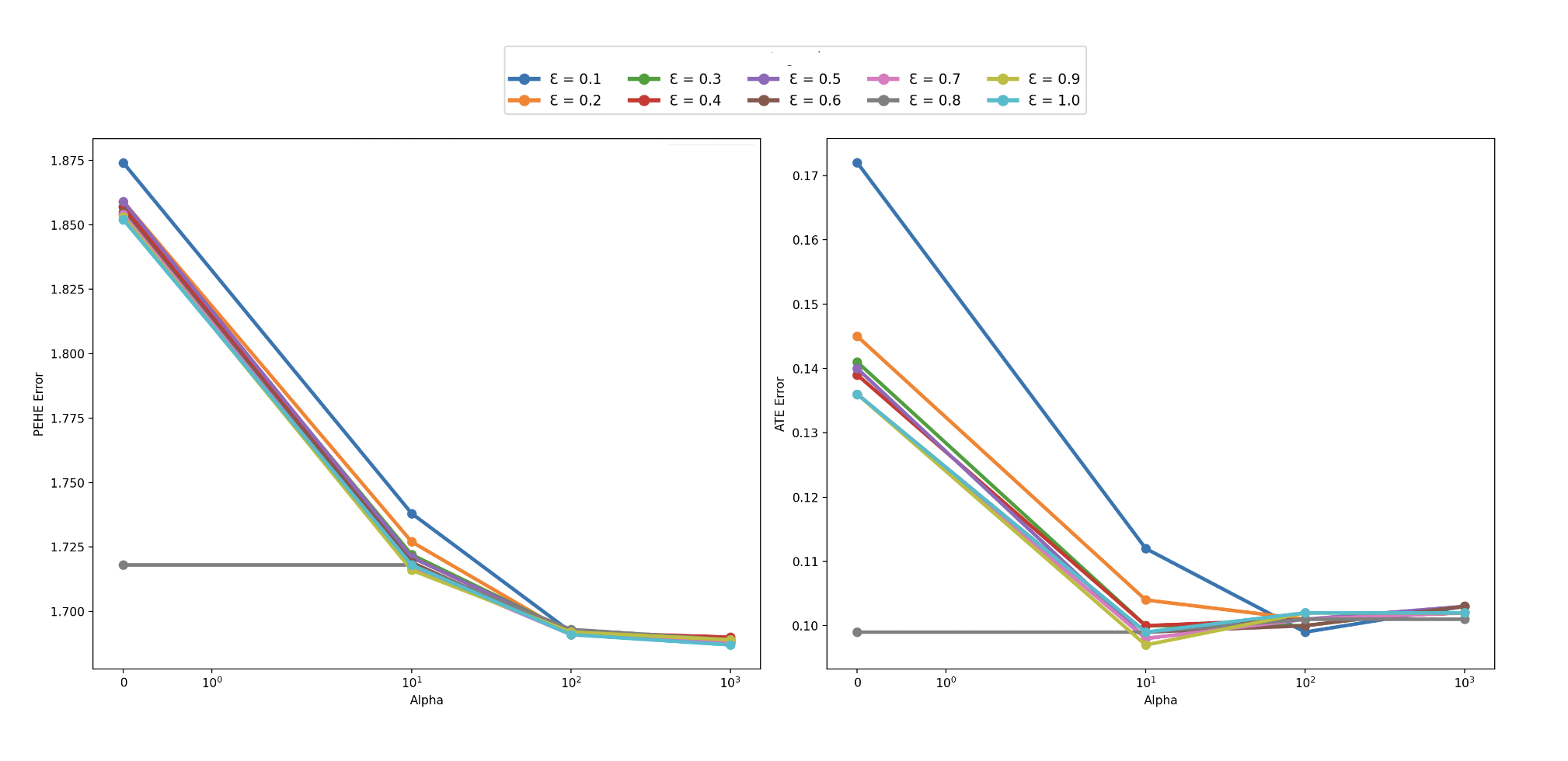}
    \caption{We showcase the ablation study performed to understand the behaviour of the IPM scaling hyperparameter $\alpha$ and the Fish update hyperparameter, $\epsilon$. As we see, even though the difference in the ATE errors is quite low, we outperform all neural baselines while having a lower PEHE error. This means that performing gradient matching is advantageous before performing covariate matching.}
    \label{fig:IHDP2}
\end{figure*}

\begingroup % localize scope of next instruction 
\begin{table*}[t] 
\setlength\tabcolsep{4pt}
\noindent
\centerline{
\begin{tabular*}{\textwidth}
{@{\extracolsep{\fill}} *{9}{c}}
\toprule 
{\multirow{1.2}{*}{\textbf{Sample Type}}} & \textbf{Model} &
\multicolumn{2}{c}{\textbf{IHDP}} & 
%\cmidrule{3-5}
\multicolumn{1}{c}{\textbf{Jobs}} &
%\cmidrule{5-7}
\multicolumn{1}{c}{\textbf{Cattaneo}} & \\
& & $\sqrt{\epsilon_{PEHE}}$ & $\epsilon_{ATE}$ &  $normalised \, \, \epsilon_{ATT}$ 
%& $\epsilon_{ATC}$  
& \small$ATE \in [-250, -200]$\\ %& $\epsilon_{ATT}$\\ 
\midrule
\addlinespace
\multirow{13.2}{*}{\footnotesize Within - Sample} & \footnotesize 

T-learner & \footnotesize $1.92  \pm  0.00$ & \footnotesize $1.06  \pm  1.26$  & \footnotesize $0.5875  \pm  0.63$  & \footnotesize $-358.20  \pm  293.74$ \\

& \footnotesize S-learner & \footnotesize $1.97  \pm  0.11$ & \footnotesize $1.30  \pm  1.09$   & \footnotesize $0.1738  \pm  0.11$ & \footnotesize $-149.06  \pm  113.81$ \\

& \footnotesize X-Learner & \footnotesize $1.70  \pm  0.02$ & \footnotesize $0.83  \pm  0.75$  & \footnotesize $0.5290  \pm  0.48$ & \footnotesize $-334.12  \pm  269.30$ \\

& \footnotesize DA-Learner & \footnotesize $1.82  \pm  0.11$ & \footnotesize $0.98  \pm  1.08$   & \footnotesize $0.6646  \pm  0.62$ & \footnotesize $-1906.68  \pm  359.79$ \\

& \footnotesize DR-Learner & \footnotesize $2.69  \pm  0.15$ & \footnotesize $2.25  \pm  0.94$ & \footnotesize $1.0000  \pm  1.19$ & \footnotesize $-185.39  \pm  157.19$ \\

& \footnotesize IRM \cite{shah2022finding} & 
\footnotesize $1.61 \pm 0.09$ & \footnotesize $0.05 \pm 0.04$  & \footnotesize $0.0000 \pm 0.13$ & \footnotesize $-226.32 \pm 11.03$ \\

& \footnotesize TARNet & 
\footnotesize $1.98 \pm 0.06$ & \footnotesize $0.05 \pm 0.04$  & \footnotesize $0.2079 \pm 0.01$ & \footnotesize $-211.61 \pm 22.28$ \\

& \footnotesize CFRNet 
& \footnotesize $1.73 \pm 0.03$ & \footnotesize $0.11 \pm 0.035$   & \footnotesize $0.2066 \pm 0.01$ & \footnotesize $-210.21 \pm 24.32$ \\

& \footnotesize Alt-M-TARNet & \footnotesize $2.00 \pm 0.06$  & \footnotesize $0.15 \pm 0.014$ & \footnotesize $0.0003 \pm 0.22$ & \footnotesize $-272.39 \pm 33.72$ \\

& \footnotesize Alt-M-CFRNet & \footnotesize $1.92 \pm 0.06$ & \footnotesize $0.08 \pm 0.003$ & \footnotesize $0.0023 \pm 0.13$ & \footnotesize $-276.43 \pm 14.32$ \\

& \footnotesize Seq-M-TARNet & \footnotesize $1.84 \pm 0.03$  & \footnotesize $0.14 \pm 0.021$ & \footnotesize $0.0344 \pm 0.005$ & \footnotesize $-221.72 \pm 20.69$ \\

& \footnotesize Seq-M-CFRNet & \footnotesize $1.65 \pm 0.00$ & \footnotesize $0.08 \pm 0.003$ & \footnotesize $0.0363 \pm 0.005$ & \footnotesize $-223.50 \pm 17.05$ \\
\addlinespace

\addlinespace
\multirow{13.2}{*}{\footnotesize Out-of - Sample} & \footnotesize

T-learner & \footnotesize $1.92  \pm  0.00$ & \footnotesize $1.06  \pm  1.26$  & \footnotesize $0.6164  \pm  0.65$  & \footnotesize $-358.20  \pm  293.74$ \\

& \footnotesize S-learner & \footnotesize $1.97  \pm  0.11$ & \footnotesize $1.30  \pm  1.09$   & \footnotesize $0.2317  \pm  0.13$ & \footnotesize $-149.06  \pm  113.81$ \\

& \footnotesize X-Learner & \footnotesize $1.70  \pm  0.02$ & \footnotesize $0.83  \pm  0.75$  & \footnotesize $0.5619  \pm  0.46$ & \footnotesize $-334.12  \pm  269.30$ \\

& \footnotesize DA-Learner & \footnotesize $1.82  \pm  0.11$ & \footnotesize $0.98  \pm  1.08$   & \footnotesize $0.6881  \pm  0.63$ & \footnotesize $-1906.68  \pm  359.79$ \\

& \footnotesize DR-Learner & \footnotesize $2.69  \pm  0.15$ & \footnotesize $2.25  \pm  0.94$ & \footnotesize $1.0000  \pm  1.23$ & \footnotesize $-185.39  \pm  157.19$ \\

& \footnotesize IRM \cite{shah2022finding} & 
\footnotesize $1.61 \pm 0.09$ & \footnotesize $0.05 \pm 0.04$  & \footnotesize $0.0700 \pm 0.009$ & \footnotesize $-226.32 \pm 11.03$ \\

& \footnotesize TARNet & 
\footnotesize $1.95 \pm 0.08$ & \footnotesize $0.007 \pm 0.003$  & \footnotesize $0.2400 \pm 0.02$ & \footnotesize $-225.61 \pm 15.01$ \\

& \footnotesize CFRNet 
& \footnotesize $1.67 \pm 0.04$ & \footnotesize $0.009 \pm 0.005$   & \footnotesize $0.2396 \pm 0.02$ & \footnotesize $-223.99 \pm 14.77$ \\

& \footnotesize Alt-M-TARNet & \footnotesize $1.96 \pm 0.078$  & \footnotesize $0.012 \pm 0.016$ & \footnotesize $0.0003 \pm 0.22$ & \footnotesize $-272.39 \pm 33.72$ \\

& \footnotesize Alt-M-CFRNet & \footnotesize $1.78 \pm 0.01$ & \footnotesize $0.003 \pm 0.002$ & \footnotesize $0.0023 \pm 0.13$ & \footnotesize $-276.43 \pm 14.32$ \\

& \footnotesize Seq-M-TARNet & \footnotesize $1.78 \pm 0.031$  & \footnotesize $0.07 \pm 0.018$ & \footnotesize $0.0000 \pm 0.32$ & \footnotesize $-232.39 \pm 12.57$ \\

& \footnotesize Seq-M-CFRNet & \footnotesize $1.57 \pm 0.001$ & \footnotesize $0.003 \pm 0.002$ & \footnotesize $0.0017 \pm 0.33$ & \footnotesize $-232.78 \pm 11.36$ \\
\addlinespace

\addlinespace
\bottomrule
\end{tabular*}}
\caption{Tabular Results across all datasets. Our method Seq-M-CFRNet surpasses all other meta-learners as well as neural methods TARNet and CFRNet. Performing gradient matching before covariate matching helps in removing spurious correlations and reducing effects of confounding.}
\label{tab:results}
\end{table*}
\endgroup

\begingroup % localize scope of next instruction 
\begin{table*}[t!]
\noindent
\centerline{
\begin{tabular*}{\textwidth}
{@{\extracolsep{\fill}} *{5}{c}}
\toprule 
\textbf{Model} &
\multicolumn{2}{c}{\textbf{Within-Sample}} & 
%\cmidrule{3-5}
\multicolumn{2}{c}{\textbf{Out-of-Sample}} \\
%\cmidrule{5-7}
 & $\epsilon_{mPEHE}$ & $\epsilon_{mATE}$ & $\epsilon_{mPEHE}$ & $\epsilon_{mATE}$\\  
\midrule
\addlinespace
            
 \footnotesize SC - TARConv & \footnotesize$0.573 \pm 0.003$ & \footnotesize$0.180 \pm 0.005$ & \footnotesize$0.578 \pm 0.003$ & \footnotesize$0.192 \pm 0.001$\\

 \footnotesize SC - CFRConv & \footnotesize$0.371 \pm 0.001$ & \footnotesize$0.101 \pm 0.001$ &  \footnotesize$0.383 \pm 0.002$ & \footnotesize$0.110 \pm 0.000$\\

  \footnotesize Seq-M-SC - TARConv & \footnotesize $0.368 \pm 0.002$ & \footnotesize $0.167 \pm 0.002$ & \footnotesize{$0.376 \pm 0.004$} & \footnotesize $0.146 \pm 0.003$\\

 \footnotesize Seq-M-SC - CFRConv  & \footnotesize{$0.365 \pm 0.000$} & \footnotesize{$0.112 \pm 0.000$} & \footnotesize{$0.368 \pm 0.001$} & \footnotesize{$0.118 \pm 0.001$}\\

\addlinespace
\bottomrule
\end{tabular*}}
\caption{Results for the estimation of the employment of various guides on the maximum evacuation time taken \cite{takeuchi2021grab}. We surpass the respective methods in terms of the PEHE errors, and perform well in estimating the ATEs simultaneously.}
\label{tab:results_images}
\end{table*}
\endgroup
\section{Experiments}\label{sec:experiments}
\subsection{Datasets}
\paragraph{Tabular Data} We perform experiments on three causal benchmark datasets, i.e., 
\begin{itemize}
    \item \textbf{IHDP} is a randomized controlled study designed to evaluate the effect of home visit from specialist doctors on the cognitive test scores of premature infants \cite{hill2011bayesian}. This dataset is semi-synthetic, and selection bias is induced by removing non-random subsets of the treated individuals to create an observational dataset, and the outcomes are generated using the original covariates and treatments. It contains 747 subjects and 25 variables. Since this is a semi-synthetic dataset, we have access to its true ATE values.  We induce confounding by dropping a few features, and define the birth-weight of the child as the anchor variable. For more details, refer to \ref{IHDP_data}.

    \item \textbf{Jobs} by LaLonde is a widely used real-world benchmark in the causal inference community, where the treatment is job training and the outcomes are income and employment status after training \cite{lalonde1986evaluating}. The dataset includes 8 covariates such as age, education, and previous earnings. This dataset combines a randomized study based on the National Supported Work program with observational data to form a larger dataset \cite{smith2005does}. In our experiments, we define the age of the person as the anchor variable.

    \item \textbf{Cattaneo} dataset studies the effect of maternal smoking on babies’ birth weight. The 20 observed features measure various attributes about the children, their mothers and their fathers \cite{cattaneo2010efficient}. The dataset considers the maternal smoking habit during pregnancy as the treatment with 864 samples and 3778 samples in the control group. This is also a real world datasets, where the counterfactual scenario is inaccessible. \cite{almond2005costs} mentions that the ATE should lie in the range (-250, 200), i.e., the effect of a Mother smoking reduces the babies’ birth weight by around ~200 grams. We induce confounding by dropping a few features and define the age of the mother as the anchor variable. For more details refer to \ref{Cattaneo_data}.

\end{itemize}
The results for these datasets can be seen in Table \ref{tab:results}.  
\paragraph{Analysis of Runtimes} Since we use a linear approximation of the Inter-Domain Gradient Matching oracle, known as the FISH algorithm as given in Shi et al, the cost of computation is minimal. Covariate matching requires only the MMD metric or Wasserstein distance in order to penalize the distributional difference, and hence is not computationally very costly. We have used an Apple Mac laptop (M2 processor) with a 10-core GPU. Our running times for the IHDP, Jobs, and the Cattaneo datasets were $1.23 \pm 0.05s$, $0.48 \pm 0.12s$, and $1.44 \pm 0.05s$ respectively.

\paragraph{Image Data} We work on the crowd management dataset showcased by \cite{takeuchi2021grab}. This work focuses on the problem of estimating the cost of crowd movement (such as evacuation times etc) from a relatively small amount of biased data. The data is modeled based on a simulated situation of crowds, such as their current locations, which are observed as covariates. These covariates can be denoted as $\mathcal{X} \in \mathbb{R}^{d}$, where $d$ is the number of people. A guide (treatment) is $Z \in \mathcal{Z} \subset \{0, 1\}^{t}$ that consists of $t$ possible actions, where $\mathcal{Z}$ is a set of possible guides. A guide can select single or multiple actions (i.e., a combination of actions); therefore, $|\mathcal{Z}|$ grows exponentially as $t$ increases. Potential outcomes are denoted as $Y_{i} \in \mathbb{R}$. We suppose that two arbitrary guide plans A and B correspond to guides $z$ and $z'$, respectively. Given covariates $\mathcal{X}$ and a guide $z$, we denote the cost consumed by crowds as $Y_{x}(z)$. The conditional average treatment effect (CATE) for covariates $\mathcal{X}$ can be defined as $\mathbb{E}[Y_{x}(z)] -[\mathbb{E}Y_{x}(z')]$. The results can be seen in Table \ref{tab:results_images}.  \label{sec:img}

\subsection{Baselines}
\paragraph{Tabular Baselines} We experiment with various existing meta-learners \cite{kunzel2019metalearners}, namely the \textit{S-Learner, T-Learner}, and the \textit{X-Learner} and include the the \textit{Domain Adaptation-Learner} and the \textit{Doubly Robust-Learner} from the EconML python library \cite{econml}  for heterogeneous treatment effect estimation.
We include neural techniques \textit{TARNet} and \textit{CFRNet} \cite{johansson2016learning, shalit2017estimating, johansson2022generalization} as our baselines. We use the implementation given by \cite{Asami}. We name our methods Alt-M-TARNet, Alt-M-CFRNet, Seq-M-TARNet, Seq-M-CFRNet as we perform alternating as well as sequential optimization between gradient matching and Treatment Agnostic or the Counterfactual Regression Networks.

\paragraph{Image-Based Baseline} The authors propose a spatial convolutional counterfactual regression (SC-CFR). The architecture consists of a few extra layers for convolution and pooling to project the set of covariates to a suitable feature space. We introduce gradient matching in this architecture to motivate the need to obtain domain-independent representations. We use the architectures showcased by the authors of the curated dataset \cite{takeuchi2021grab} coined \textit{SC-TARConv} and \textit{SC-CFRConv} (With zero and non-zero IPM terms respectively), which contains extra spatial convolutional layers, along with pooling. We create domain based representations for this in a similar way as we did for tabular data, and coin our method \textit{Seq-M-SC-TARConv} and \textit{Seq-M-SC-CFRConv} respectively.

\subsection{Metrics}
\paragraph{Tabular Metrics} We have the following metrics for tabular experiments:
\begin{itemize}
    \item ATE: Measures the difference in mean (average) outcomes between units assigned to the treatment and units assigned to the control: 
    \begin{align}
        ATE = \mathbb{E}_{x \sim p(x)}[\tau(x)]
    \end{align}
    Where, $\tau(x) := \mathbb{E}[Y_{1} - Y_{0} | x]$, also called the Individual Treatment effect (ITE). Due to the semi-synthetic nature of the IHDP dataset, we have access to the ITEs and hence the actual ground truth. Hence, we can report the error in ATE estimation as shown in Table \ref{tab:results}.

    \item PEHE: The error in Precision in Estimating Heterogeneous Effects, $\epsilon_{PEHE}(f)$ as defined by \cite{shalit2017estimating}, is a measure of the uncertainty in the predictions, where $f : \mathcal{X} \times \{0, 1\} \longrightarrow \mathcal{Y}$. In the case of semi-synthetic datasets, since we have access to the ground truth, we define this metric as: 
    \begin{align}
        \epsilon_{PEHE}(f) = \int_{\mathcal{X}}(\hat{\tau_{f}(x)} - \tau_{f}(x))^{2}p(x)dx
    \end{align}

    \item ATT and ATC: For datasets where we do not have access to the ground truths we have access to the ground truths for only the treated individuals, as these are factuals. In the case of the jobs dataset, the true ATT lies around \textit{\small{1676.3426}}, and hence we can calculate the ATT error. For the ATC, we can compute the true average treatment on the control by, 
    \begin{align}
        ATC = |C|^{-1} \Sigma_{i \in C}(y_i) - |T \cap E|^{-1} \Sigma_{i \in (T \cap E)}(y_i)
    \end{align}
    Since we are interested in the differences in the factual and the counterfactual scenario, we evaluate a threshold, $\mathcal{T}$, as: 
    \begin{align}
        \mathcal{T} = \frac{1}{|C|}\Sigma_{i \in C}(f(x_{i}, 1) - f(x_{i}, 0))
    \end{align}
    The ATC error can be calculated by the difference in the calculated value and the threshold, $\mathcal{T}$. We report the normalised scaled ATT and ATC errors in our tabular results.
\end{itemize}

\paragraph{Image-Based Metrics} As defined by \cite{takeuchi2021grab}, since the data contains a set of treatments which determine the guide, we employ the $\epsilon_{mATE}$ and $\epsilon_{mPEHE}$ to showcase our results. They are respectively defined as,
\begin{align}
    \epsilon_{mATE} = \frac{1}{\binom{t}{2}}\Sigma_{(z_i, z_j) \in \mathcal{Z}}(\epsilon_{ATE})
\end{align}
\begin{align}
    \epsilon_{mPEHE} = \frac{1}{\binom{t}{2}}\Sigma_{(z_i, z_j) \in \mathcal{Z}}(\epsilon_{PEHE})
\end{align}
We showcase these metric in our results in Table \ref{tab:results_images} along with the Root mean square error.

\subsection{Results and Discussion}
\paragraph{Tabular Results}\label{sec:tab_res_disc}
As we see in Table \ref{tab:results}, our proposed neural architectures reduce the ATE errors considerably for the IHDP and the Cattaneo datasets. However, performance on the jobs dataset is average, mainly due to its size. In terms of the model uncertainty, robust meta-learners like X-learner and the Domain-Adaptation learner \cite{kunzel2019metalearners} also perform fairly well in lowering PEHE errors. In the case of the IHDP dataset, the true ATE is known and lies close to the value \textit{4.016}. Our method gives the lowest value for the ATE and PEHE errors. We are interested in its generalisability and hence are interested in minimizing the out-of-sample errors. Our method, Seq-M-CFRNet performs the best with an error of only \textit{0.003 $\pm$ 0.002} surpassing all other methods. In the case of the other two datasets too, our method got \textit{0.0017 $\pm$ 0.03} as the normalised ATE error, and estimated the ATE for the Cattaneo dataset in the correct range, i.e., \textit{-232.78 $\pm$ 11.36}. \textit{Here, usage of only either FISH (IDGM) or CFRNet (Covariate Matching) is surpassed by combining them and sequentially minimizing their errors as in our framework.} There is also a trade-off between various update hyperparameters $\epsilon$ (IDGM parameter) and $\alpha$ (covariate matching parameter). Our method combines the IRM proxy that ensures ignorable feature finding with covariate matching. When the first step works exactly (achieving exact invariance) the resulting representation’s quality is equivalent to that of the CFRNet. Sequentiality ensures this. While we reported an alternating minimization based implementation, it is not obvious if the resulting representation is good with respect to either criteria. Further the IRM proxy solves a complicated bi-level objective and covariate matching could be described by an extra MMD-based loss function. So this is not a usual multi-objective problem where one could combine losses to give some sort of pareto optimality. When the first stage is approximate, we should use our theorem that approximates conditional independence, when it holds in an SEM, implying an interval bound on the treatment effect justifying making small errors in the first step. We also experimented with alternating optimization of the batch-wise IDGM followed by Covariate Matching constraint imposition. However, this approach resulted in higher ATE and PEHE errors. This implies that obtaining the ignorable covariates before performing covariate matching is giving better results than simultaneously obtaining gradient and covariate matched representations.
% \begin{itemize}
%     \item IHDP Dataset: Alternating minimization gave PEHE and ATE errors as $1.89 \pm 0.06$ and $0.08 \pm 0.006$, while sequential minimization gave $1.65 \pm 0.00$ and $0.003 \pm 0.002$ 
%     \item Jobs dataset: Both approaches performed similarly.
%     \item Cattaneo: Alternating minimization resulted in ATE value $-276.43 \pm 14.32$, while sequential minimization gave $-232.78 \pm 11.36$ where the ideal range was between $(-250,-200)$. 
% \end{itemize}
\paragraph{Image Results} As shown in Table \ref{tab:results_images}, SC-TARConv performed quite poorly as compared to SC-CFRConv and both of our variants \cite{takeuchi2021grab}. We surpass both baseline models in terms of the PEHE errors, while having a much lower ATE error than SC-TARConv and almost matching it in the case of SC-CFRConv. It is safe to assume that performing inter-domain gradient matching before performing covariate matching to isolate the causal effect works for multi-modal settings. However, choosing an anchor variable, and ensuring that the domains lie within the \textit{linear general position} \cite{arjovsky2019invariant} are two factors need to be taken care of.

\section{Conclusions and Future Work}

Estimating treatment effects from observational data is challenging as the control and treatment groups do not have identical distributions and have hidden and unmeasured confounding. We propose a novel neural net architecture that yields a representation that is a valid adjustment in the presence of confounding and simultaneously ensures that covariate distribution match across treatment groups.
Our method is robust, interpretable, and explainable as all other neural alternatives like CFRnet.

We show empirically that approximate inter-domain gradient matching (an IRM oracle) followed by covariate matching based representation regularization can significantly improve performance on causal benchmark datasets. Our ATE and PEHE errors are significantly lower across tabular datasets and we match the ATE errors in our image dataset while lowering the PEHE errors extensively. We also theoretically justify usage of approximate IRM oracle and covariate matching techniques in terms of causal effect efect isolation in Appendix \ref{sec:proof}. However, obtaining interval bounds for treatment effect is an interesting future direction.

% \begin{contributions} % will be removed in pdf for initial submission 
% 					  % (without ‘accepted’ option in \documentclass)
%                       % so you can already fill it to test with the
%                       % ‘accepted’ class option
%     Briefly list author contributions. 
%     This is a nice way of making clear who did what and to give proper credit.
%     This section is optional.

%     H.~Q.~Bovik conceived the idea and wrote the paper.
%     Coauthor One created the code.
%     Coauthor Two created the figures.
% \end{contributions}

% \begin{acknowledgements} % will be removed in pdf for initial submission,
% 						 % (without ‘accepted’ option in \documentclass)
%                          % so you can already fill it to test with the
%                          % ‘accepted’ class option
%     Briefly acknowledge people and organizations here.

%     \emph{All} acknowledgements go in this section.
% \end{acknowledgements}

% References

\subsubsection*{Broader Impact Statement}
Our method is highly replicable and has real-world applications, making it valuable to domain experts, particularly in healthcare and biomedicine. However, it is crucial to acknowledge the potential risks of misuse, especially in biased settings without proper fairness assessments. To ensure ethical deployment, our approach should be applied with caution in high-stakes domains, prioritizing fairness and transparency.

\newpage
\bibliography{bibliography}
\onecolumn
\newpage
\title{Appendix}
\maketitle
\section{Proof for Theorem \ref{thrm:our_theorem}}\label{sec:proof}
\begin{proof}  
We are given a linear structural equation model (SEM) where, $\mathcal{V}$ denotes the set of all variables containing hidden confounders $\mathcal{U}$, covariates $\mathcal{X}$, treatment variable $T$ and outcome $Y$. All these variables are jointly Gaussian and have zero mean. Consider any $X_t \in \mathcal{X}$, that has a direct edge to $T$ called the anchor variable. Let $E$ be sub-sampled using $X_t$, i.e., $e = f(X_t, \eta)$  where $\eta$ is a noise variable independent of the number of vertices in the graph, and $f$ is a function of $X_t$ (realization of $X_t$) and $\eta$. Let $\mathcal{X}^{-X_t}$ denote all features without $X_t$ and its realization is denoted $x^{-X_t}$. We also assume that $T$ has only one child, $Y$. Further, we also assume that $Y$ has no children. 
This implies we can express the SEM as:
\begin{equation*}
    \mathcal{V} = (\mathcal{V}_1, ..., \mathcal{V}_p)^{T} 
\end{equation*}
\begin{equation}
    \mathcal{V} = B\mathcal{V} + \varepsilon
\end{equation}
Here, $B$ denotes the edge weights matrix such that, $B \in \mathbb{R}^{p \times p}$, and $B_{i,j} \neq 0$ iff $\mathcal{V}_i \longrightarrow \mathcal{V}_j \in E$. $\varepsilon \in \mathbb{R}^{p \times 1}$, is a multi-variate Gaussian distributed random variable with zero mean and whose covariance matrix, $\Omega := diag(\sigma_1 ^2, .. ,\sigma_p ^2)$.
\begin{itemize}
    \item We are given $\epsilon$-approximate invariance, i.e, $cov(Y, X_t |T, Z) < \epsilon \,\, (\text{for some } \epsilon > 0)$,
    \item \label{comment 1} We assume $X_t$ is highly correlated to $T = \alpha X_t + \beta [\mathcal{V} \backslash X_t] + \varepsilon_T,\, \text{such that \,} \{\alpha \rightarrow 1, \beta \rightarrow 0\} $ and $\mathbb{E}[\varepsilon_T] = 0$.   
\end{itemize}

\begin{quote}
    \textit{The goal is to obtain an upper bound for the covariance of $Y'$ (The outcome $Y$ after intervention on $T$) and $T$ given some set $Z$, as this would imply approximate causal effect isolation. This would imply the set $Z$ is an approximate valid adjustment set i.e.,
    $$cov(Y', T |Z) \, \text{\textit{is bounded}}$$
    Since we assume that the SEM is linear, showing that $cov(Y', X_t |T, Z) < \epsilon$ would imply that $cov(Y', T | Z) < \epsilon$.}
\end{quote}

\begin{figure*}
    \centering
    \includegraphics[width = 0.85\linewidth]{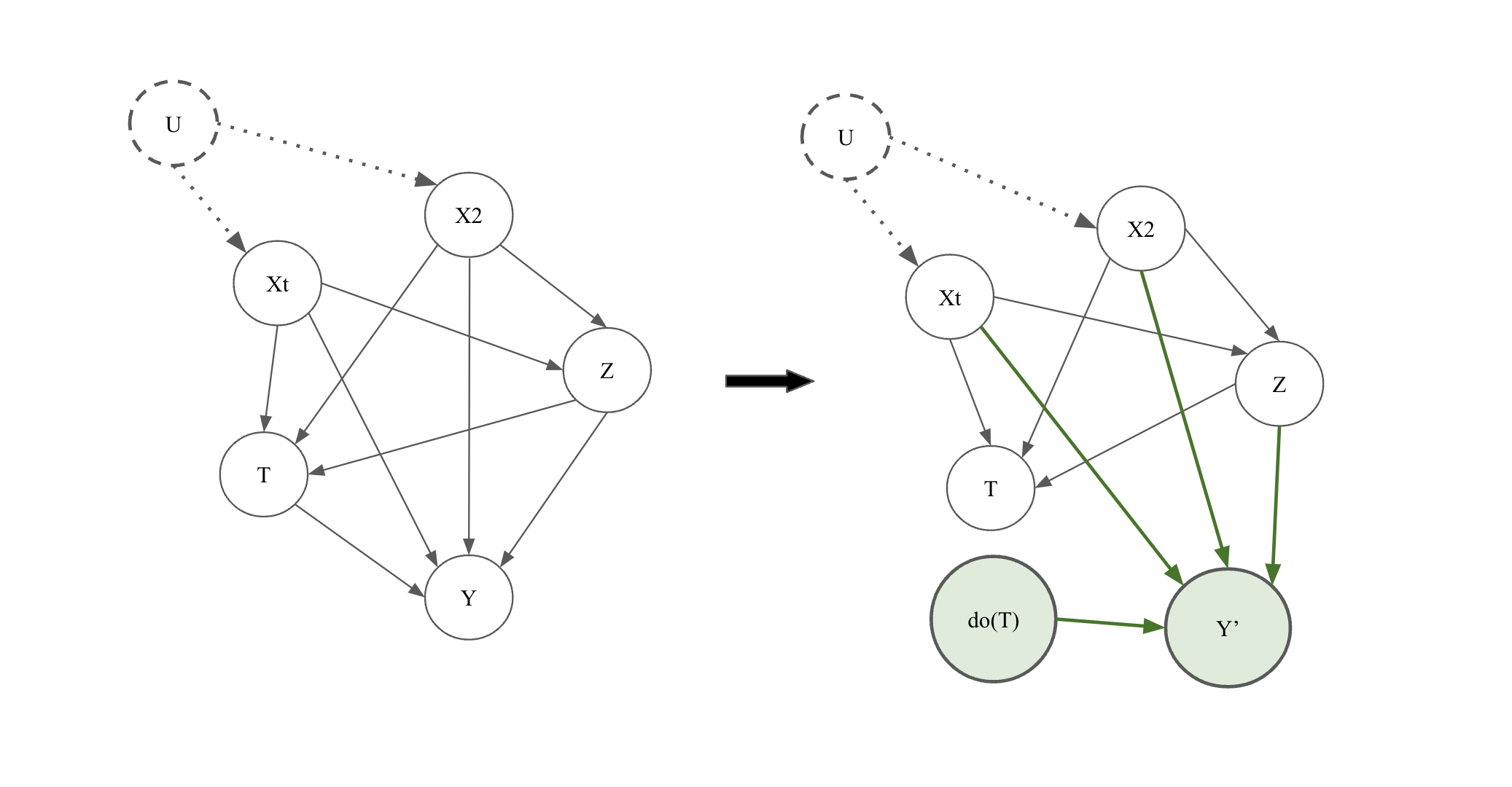}
    \caption{An example of a causal graph pre-intervention and post-intervention.}
    \label{fig:proof_pic}
\end{figure*}

\paragraph{Note 1} \label{Note 1} Pre-intervention, $T$ contains atleast one incoming edge, and has exactly one outgoing edge. \\

According to the path variables formula given in \cite{uhler2013geometry}, the conditional covariance between two variables $Y$ and $X_t$ given the subset $\{T \cup Z\}$ can be expressed as the product of weights from $B$ on self-avoiding paths lying between $Y$ and $X_t$ which are unblocked while variables $\{T \cup Z\}$ are conditioned upon, as follows:  

\begin{equation}
    P_{Y, X_t | T, Z} = \text{det}(K_{Q^cQ^c}) K_{Y, X_t |T, Z} - K_{Y, Q^c} C(K_{Q^c Q^c}) K_{Q^cX_t} < \, \epsilon \,  (\text{given})
\end{equation}

where, $K = I-B$, $Q = \{Y, X_t\} \cup \{T,Z\}$, $Q^c = \mathcal{V} \backslash \{Y, X_t, T, Z\}$

\paragraph{} Now, if we intervene on the variable T, we add a new variable $do(T)$ to the system. This new variable only has an edge outgoing to $Y$, and the edge from $T$ to $Y$ is now removed. An example is showcased in Fig \ref{fig:proof_pic}. Since the distribution of $Y$ changes, we now refer to it as $Y'$.

\paragraph{Note 2} $T$ in the post-interventional SEM is a collider, i.e., it only has incoming edges.
 
Hence, we can now say,
\begin{equation} \label{eq: paths_post}
    P_{Y', X_t | T, Z} = \text{det}(K'_{Q^cQ^c}) K'_{Y', X_t |T, Z} - K'_{Y', Q^c} C(K'_{Q^c Q^c}) K'_{Q^cX_t} 
\end{equation}
where, $K' = I-B'$, while $Q = \{Y', X_t\} \cup \{T, Z\}$ and $Q^c = \mathcal{V} \backslash \{Y', X_t, T, Z\}$. 
\paragraph{} Hence, intervening on $T$ has effectively split it into two variables, $T$ and $do(T)$ (Fig \ref{fig:proof_pic}), and since $do(T)$ is a terminal node (as it has no incoming edges) and $T$ a collider, the number of paths that are unblocked in $P_{Y, X_t | T, Z} > P_{Y', X_t | T, Z}$. This implies that the association flowing from $Y$ to $X_t$ (pre-intervention) must be more than that between $Y'$ and $X_t$ (post-intervention);
\begin{equation}
    \implies cov(Y', X_t | T, Z) = \epsilon' < cov(Y, X_t | T, Z) < \epsilon
\end{equation}

\paragraph{Note 3} $B' = A + A^T - AA^T$ (Both $A, B$ are symmetric and positive semi-definite).\\

Let us first express $P_{Y', X_t |T, Z}$ in terms of polynomials of $B'$ using Ponstein's Theorem as given in \cite{uhler2013geometry}:

\begin{align*}
    & \underbrace{\left[1 + \sum_{k=1}^{|Q^c|} \sum_{\substack{m_1 + \cdots + m_s = k}} (-1)^s \mu(c_{m_1}) \cdots \mu(c_{m_s}) \right]}_{\text{Term 1.1}} 
    \underbrace{\left[\sum_{k:Y' \to k \leftarrow X_t} a_{Y'k} a_{k X_t} - a_{Y' X_t}\right]}_{\text{Term 1.2}} \\
    & - \underbrace{\left[\sum_{q \in Q^c} \sum_{k:Y' \to k \leftarrow q} a_{Y'k} a_{k q} - a_{Y' q} \right]}_{\text{Term 1.3}}
    \underbrace{\left[\sum_{\substack{i, j \in Q^c \\ i \neq j}} \sum_{k=2}^{|Q^c|} 
    \sum_{\substack{m_0 + \cdots + m_s = k-1}} (-1)^s \mu(d_{m_0}) \mu(c_{m_1}) \cdots \mu(c_{m_s})\right]}_{\text{Term 1.4}} \\
    & \underbrace{\quad \times \left[\sum_{q \in Q^c} \sum_{k:q \to k \leftarrow X_t} a_{qk} a_{k X_t} - a_{q X_t}\right]}_{\text{Term 1.5}} < \epsilon'
\end{align*}

Here, $\mu(d_{m_0})$ denotes the product of the edge weights along a self-avoiding path of length $m_0$ from $Y'$ to $X_t$ in the causal subgraph restricted only to nodes in $Q^c$. $\mu(c_{m_0}).. \mu(c_{m_s})$ denote the product of the edge weights along self-avoiding cycles in the causal subgraph restricted only to nodes in $Q^c$ of lengths $m_1,..,m_s$ respectively. Here, $d_{m_0}, c_{m_1}, .., c_{m_s}$ are all disjoint paths.

Let us now consider a different set for conditioning, i.e.,
\begin{equation} \label{eq: cond_z} 
    P_{Y', X_t | Z} = \text{det}(K'_{Q'^cQ'^c}) K'_{Y', X_t |Z} - K'_{Y', Q'^c} C(K'_{Q'^c Q'^c}) K'_{Q'^cX_t} 
\end{equation}
Here, K remains the same, but $Q' = \{Y', X_t\} \cup \{Z\}$, $Q'^c = \mathcal{V} \backslash \{Y', X_t, Z\}$, which implies $Q'^c = Q^c \cup \{T\} \,\, , \,\, |Q'^c| = |Q^c| +1$. Hence $P_{Y', X_t |Z}$ would equate to

\begin{align*}
    & \underbrace{\left[1 + \sum_{k=1}^{|Q^c| + 1} \sum_{\substack{m_1 + \cdots + m_s = k}} (-1)^s \mu(c_{m_1}) \cdots \mu(c_{m_s}) \right]}_{\text{Term 2.1}}
    \underbrace{\left[\sum_{k:Y' \to k \leftarrow X_t} a_{Y'k} a_{k X_t} - a_{Y' X_t}\right]}_{\text{Term 2.2}} \\
    & - \underbrace{\left[\sum_{q \in Q^c \cup \{T\}} \sum_{k:Y' \to k \leftarrow q} a_{Y'k} a_{k q} - a_{Y' q} \right]}_{\text{Term 2.3}}
    \underbrace{\left[\sum_{\substack{i, j \in Q^c \cup \{T\} \\ i \neq j}} \sum_{k=2}^{|Q^c|+1} 
    \sum_{\substack{m_0 + \cdots + m_s = k-1}} (-1)^s \mu(d_{m_0}) \mu(c_{m_1}) \cdots \mu(c_{m_s})\right]}_{\text{Term 2.4}} \\
     & \underbrace{\quad \times \left[\sum_{q \in Q^c \cup \{T\} } \sum_{k:q \to k \leftarrow X_t} a_{qk} a_{k X_t} - a_{q X_t}\right]}_{\text{Term 2.5}} 
\end{align*}

\paragraph{Note 4} Let $0 < \text{det}(K'_{Q^cQ^c}) K'_{Y', X_t |T, Z} <  \epsilon '$. 
\begin{quote}
    \textit{This ensures the fact that the terms constituting the path formula are also small, and that the big terms do not cancel out.}
\end{quote}

Using this final assumption, let us compare all the terms in both the expressions, as shown above individually:
\begin{itemize}
    \item For term 1.1, note that $\{Y', X_t, T, Z\} \notin Q^c$. For Term 2.1, $Q'^c = Q^c \cup \{T\}$. However, since $T$ is a collider and is not conditioned on, no backdoors via $T$ will exist. Hence the paths in consideration across both terms, are the same. Since the expressions in consideration are the summation of the products of edge weights along self-avoiding cycles induced by the respective subgraphs, $\text{Term 1.1} = \text{Term 2.1}$.
    
    \item Since we consider all paths from all $k: Y' \rightarrow k \leftarrow X_t$ in the entire causal graph, the former terms will contain a greater number of paths, as it conditions over $T$ (collider), i.e., unblocks paths via $T$. The second term ensures that the paths via $T$ are blocked due to the absence of conditioning on it. Since $A$ across both cases is the same except for the row corresponding to $T$, and the number of paths in consideration in Term 1.2 are atleast greater than the number of paths in Term 2.2, Term 1.2 $\geq$ Term 2.2.

    \begin{equation} \label{eq : pt1}
        \implies \text{det}(K'_{Q^cQ^c}) K'_{Y', X_t |T, Z} \geq \text{det}(K'_{Q'^cQ'^c}) K'_{Y', X_t |Z}
    \end{equation}
    
    \item  Term 1.3 considers paths $\{k: Y' \rightarrow k \leftarrow q, \,\, \forall q \in Q^c\}$ and Term 2.3 considers all paths $\{k: Y' \rightarrow k \leftarrow q, \,\, \forall q \in Q'^c\}$. Since $T \notin Q^c$, and paths through $T$ are blocked in Term 2.3, Term 1.3 $=$ Term 2.3. Similarly, Term 1.5 $=$ Term 2.5.
    
    \item  Comparing Terms 1.4 and 2.4, we can see that Term 2.4 contains the following extra term: \[\cancelto{0}
    {\sum_{\substack{i, j \in Q^c \\ i = T \bigoplus j = T}} \sum_{k = 2}^{|Q^c| +1} \sum_{m_0+..+m_s = k-1}(-1)^s \mu(d_{m_0})\mu(c_{m_1})..\mu(c_{m_s}) }\] \\
    Since paths through $T$ are blocked in Term 2.4, this expression equates to 0. Hence Term 1.4 $=$ Term 2.4.

\end{itemize}

\begin{equation} \label{eq : pt2}
    \implies K'_{Y', Q^c} C(K'_{Q^c Q^c}) K'_{Q^cX_t} =  K'_{Y', Q'^c} C(K'_{Q'^c Q'^c}) K'_{Q'^cX_t}    
\end{equation}

From Equations \ref{eq : pt1} and \ref{eq : pt2}
\begin{equation*}
     \implies P_{Y', X_t | Z} \leq  P_{Y', X_t | T, Z}
\end{equation*}

Hence, we can see that 
\begin{equation}
    cov(Y', X_t | Z) \leq cov(Y', X_t | T, Z)
\end{equation}

Since we know that $T$ and $X_t$ are highly correlated,

\begin{equation*}
    cov(Y', T | Z) \leq cov(Y', X_t | Z) \leq cov(Y', X_t | T, Z)
\end{equation*}
\begin{equation}
    \implies cov(Y', T |Z) \leq \epsilon' < \epsilon
\end{equation}

\end{proof}

\section{Datasets}
\subsection*{IHDP Dataset-- \cite{hill2011bayesian}}\label{IHDP_data}
We induce confounding in our data by dropping some of the variables provided in the dataset as by doing so, we lose information about our covariates on which the outcome depends. Next, to perform inter-domain gradient matching, since there is no pre-existing notion of domains in the data, we keep the feature child’s birth-weight as our anchor variable and we use it to generate environments (fake synthetic variables \cite{shah2022finding} as shown in Algorithm \ref{algo_dom_generation}). The covariates we include in our experiments contain the number of weeks pre-term that the child, the child’s head circumference at birth, birth order, neo-natal health index, mother’s age when she gave birth to the child , child’s gender, indicator for whether the mother used drugs when she was pregnant, indicator for whether the mom received any prenatal care, and the site indicator as our observed variables. The treated group consists of those infants which were provided with both intensive high-quality childcare and specialist home visits. This acts as the treatment variable.

\subsection*{Cattaneo Dataset -- \cite{cattaneo2010efficient}}\label{Cattaneo_data}
 This dataset studies the effect of maternal smoking on babies’ birth weight. The treatment variable is a binary indicator of the mother being a smoker. The feature set that we retain while inducing confounding comprises of the following attributes: mother’s marital status, indicator for whether the mother consumed alcohol when pregnant, binary variable depicting whether the child is first-born or not, mother’s age, mother’s education, mother’s race, father’s age, father’s education, father’s race, months since last birth by the mother, birth month, indicator for whether the baby is first-born, total number of prenatal care visits, number of prenatal care visits in the first trimester, and the number of trimesters the mother received any prenatal care. Apart from these, there are also a few other features available in this dataset for which we did not have access to their description. We use the age of the mother as our anchor variable to generate domains.

\subsection*{Crowd Management Dataset -- \cite{takeuchi2021grab}}
\begin{figure*}[ht]
    \centering
    \includegraphics[width = 16cm]{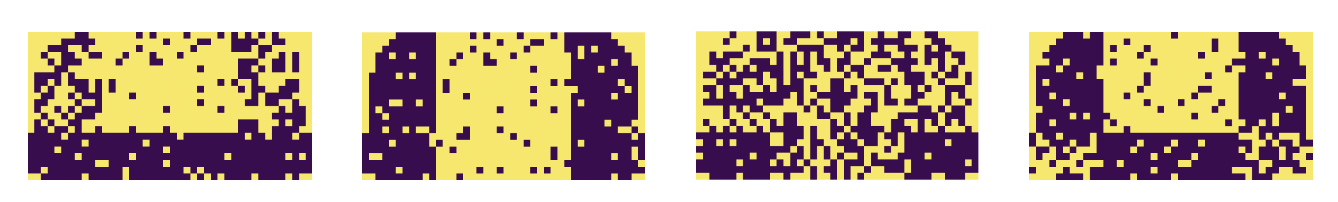}
    \caption{Example of covariates from the dataset showcased in \ref{sec:img} by \cite{takeuchi2021grab}}\label{fig:covariates}
\end{figure*}

\begin{figure}[t]
    \centering
    \includegraphics[width = 16cm]{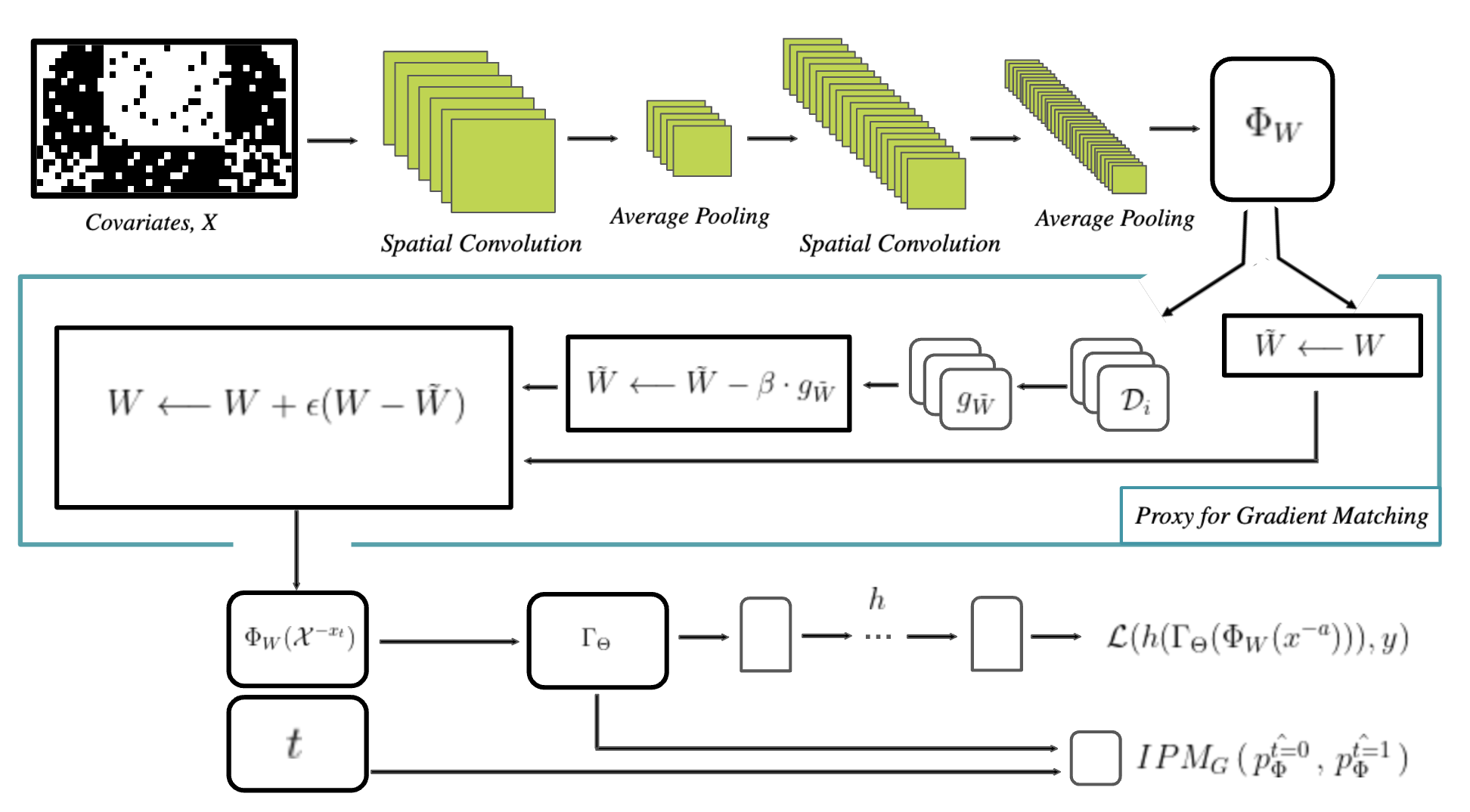}
    \caption{SC-CFR architecture, with spatial convolutional layers and pooling for image processing along with our proxy for gradient matching. We sequentially apply convolutions and pooling layers followed by gradient matching and covariate matching as shown in Figure \ref{fig:ourarch}.}
    \label{fig:ourarch_img}
\end{figure}

This dataset contains information about various people in an evacuation scenario, where the guidance provided by guides is regarded as the treatment variable. The problem under consideration is of estimating the effects of crowd movement guidance from past data. The covariates are the locations of each agent within the crowd that needs to be evacuated. They are stored as shown in Figure \ref{fig:covariates}. The covariates are within a space $\mathbb{R}^d$ where $d$ is the number of agents to be evacuated. The evacuation plan suggested by various `guides' is encoded as the treatment variable, $z = \{0,1\}^{7}$. Here, the first dimension represents the action of the guide followed by the remaining 6 values to indicate the action of the 6 exit doors. This is a non-binary treatment variable. We use the total number of agents to be evacuated as our anchor variable and generate our domains to perform IDGM. The outcome is the total evacuation time. We showcase our architecture in Figure \ref{fig:ourarch_img}. Further information on the dataset and experiments can be found in \cite{takeuchi2021grab}.

\vspace{-8pt}
\section{Mathematical Preliminaries}

\subsection{Domain generation} \label{sec:dom_gen}
Many datasets do not contain the notion of domains inherently. In order to overcome this shortcoming, we generate domains within our dataset using a procedure which authors in \cite{shah2022finding} use to determine whether a subset of covariates that don't involve the anchor variable pass a suitable conditional independence criteria for it to be deemed a valid backdoor. In our case, this anchor variable is required to be a known variable, while simultaneously should be a suitable parent of the treatment variable, $t$. This is minimal assumption which can be used to obtain the causal effect. In our case, we use the anchor variable to generate synthetic environments, which can be used as domains (Algorithm \ref{algo_dom_generation}). We ensure that the domains lie in the \textit{linear general position} \cite{arjovsky2019invariant} -- which implies that enough information is provided across domains to reduce redundancy.
\begin{algorithm}
    \caption{Domain Generation using the anchor variable \cite{shah2022finding}}\label{algo_dom_generation}
    \begin{algorithmic}
        \STATE  \textbf{Input: } Factual dataset, $(x_1, t_1, y_1), ... , (x_n, t_n, y_n)$, and a known parent variable of $t$, which we denote $X_t$. We are also given Number of environments to be generated, $m$.
        \vspace{6pt}
        \STATE $\theta^{(1)} := Uniform(1,2)$ , $\theta^{(2)} := [0]_{2 \times 2}$ ,  $\theta^{(3)} :=  Uniform(1,2)$
        \vspace{6pt}
        \STATE $\theta := (\theta^{(1)}, \theta^{(2)}, \theta^{(3)}) \in \mathbf{R}^{3}$
        \vspace{6pt}
        \RETURN Domains, $e$ $\sim$ $Softmax(\theta (X_t - \mathbf{E}[X_t])$  
    \end{algorithmic}
\end{algorithm}

\subsection{d-Separation}
d-separation is a criterion for deciding, from a given a causal graph, whether a set X of variables is independent of another set Y, given a third set Z. The idea is to associate "dependence" with "connectedness" (i.e., the existence of a connecting path) and "independence" with "unconnected-ness" or "separation". We use the terms "d-separated" and "d-connected" (d connotes "directional").\cite{pearl2009causality, shah2022finding}
\begin{quote}
    \textit{For any variables $v_{1}$, $v_{2}$ $\in \mathbf{W}$, and a set $v \subseteq \mathbf{W}$, $v_{1}$ and $v_{2}$ are d-separated by $v$ in $\mathbf{G}$ if $\mathbf{v}$ blocks every path between  $v_{1}$ and $v_{2}$ in $\mathbf{G}$.}
\end{quote}

\begingroup % localize scope of next instruction 
\begin{table*}[t]
\noindent
\centerline{
\begin{tabular*}{\textwidth}
{@{\extracolsep{\fill}} *{6}{c}}
\toprule Fish Update \\
Hyperparameter ($\epsilon$)
 & $\alpha = 0$ (TAR) & $\alpha = 10$ & $\alpha = 100$ & $\alpha = 1000$ & $\alpha = 10000$\\  
\midrule
\addlinespace
            
 \footnotesize $\epsilon = 0.1$ & \footnotesize$0.172 \pm 0.092$ & 
 \footnotesize$0.112 \pm 0.084$ & \footnotesize$0.099 \pm 0.085$ & \footnotesize$0.103 \pm 0.093$ & 
 \footnotesize$0.113 \pm 0.145$\\

 \footnotesize $\epsilon = 0.2$ & \footnotesize$0.145 \pm 0.086$ & \footnotesize$0.104 \pm 0.075$ & \footnotesize$0.101 \pm 0.085$ & \footnotesize$0.102 \pm 0.091$ & 
 \footnotesize$0.111 \pm 0.141$\\

 \footnotesize $\epsilon = 0.3$ & \footnotesize$0.141 \pm 0.085$ & \footnotesize$0.100 \pm 0.074$ & \footnotesize$0.100 \pm 0.085$ & \footnotesize$0.103 \pm 0.094$ & 
 \footnotesize$0.113 \pm 0.151$\\

 \footnotesize $\epsilon = 0.4$ & \footnotesize$0.139 \pm 0.086$ & \footnotesize$0.100 \pm 0.072$ & \footnotesize$0.101 \pm 0.085$ & \footnotesize$0.102 \pm 0.093$ & 
 \footnotesize$0.111 \pm 0.144$\\

 \footnotesize $\epsilon = 0.5$ & \footnotesize$0.140 \pm 0.086$ & \footnotesize$0.098 \pm 0.072$ & \footnotesize$0.101 \pm 0.086$ & \footnotesize$0.103 \pm 0.092$ & 
 \footnotesize$0.109 \pm 0.136$\\

 \footnotesize $\epsilon = 0.6$ & \footnotesize$0.136 \pm 0.086$ & \footnotesize$0.099 \pm 0.072$ & \footnotesize$0.100 \pm 0.086$ & \footnotesize$0.103 \pm 0.094$ & 
 \footnotesize$0.110 \pm 0.138$\\

 \footnotesize $\epsilon = 0.7$ & \footnotesize$0.136 \pm 0.085$ & \footnotesize$0.098 \pm 0.071$ & \footnotesize$0.101 \pm 0.085$ & \footnotesize$0.102 \pm 0.092$ & 
 \footnotesize$0.110 \pm 0.136$\\

 \footnotesize $\epsilon = 0.8$ & \footnotesize$0.099 \pm 0.074$ & \footnotesize$0.099 \pm 0.071$ & \footnotesize$0.101 \pm 0.085$ & \footnotesize$0.101 \pm 0.092$ & 
 \footnotesize$0.111 \pm 0.148$\\

 \footnotesize $\epsilon = 0.9$ & \footnotesize$0.136 \pm 0.085$ & \footnotesize$0.097 \pm 0.069$ & \footnotesize$0.102 \pm 0.086$ & \footnotesize$0.102 \pm 0.093$ & 
 \footnotesize$ 0.111 \pm 0.143 $\\

 \footnotesize $\epsilon = 1.0$ & \footnotesize$0.136 \pm 0.085$ & \footnotesize$0.099 \pm 0.071$ & \footnotesize$0.102 \pm 0.087$ & \footnotesize$0.102 \pm 0.091$ & 
 \footnotesize$0.111 \pm 0.147$\\

\addlinespace
\bottomrule
\end{tabular*}}
\caption{Ablation study of the ATE Errors for the IHDP dataset. The values in bold denote the least error across values for $\epsilon$ for a specific value of  $\alpha$.}
\label{tab:ablationihdptable}
\end{table*}
\endgroup
\begingroup % localize scope of next instruction 
\begin{table*}[t!]
\noindent
\centerline{
\begin{tabular*}{\textwidth}
{@{\extracolsep{\fill}} *{6}{c}}
\toprule Fish Update \\
Hyperparameter ($\epsilon$)
 & $\alpha = 0$ (TAR) & $\alpha = 10$ & $\alpha = 100$ & $\alpha = 1000$ & $\alpha = 10000$\\  
\midrule
\addlinespace
            
 \footnotesize $\epsilon = 0.1$ & \footnotesize$1.874 \pm 0.072$ & 
 \footnotesize$1.738 \pm 0.107$ & \footnotesize$1.692 \pm 0.046$ & \footnotesize$1.688 \pm 0.046$ & 
 \footnotesize$1.685 \pm 0.047$\\

 \footnotesize $\epsilon = 0.2$ & \footnotesize$1.859 \pm 0.069$ & \footnotesize$1.727 \pm 0.094$ & \footnotesize$1.691 \pm 0.046$ & \footnotesize$1.688 \pm 0.045$ & 
 \footnotesize$1.684 \pm 0.044$\\

 \footnotesize $\epsilon = 0.3$ & \footnotesize$1.857 \pm 0.070$ & \footnotesize$1.722 \pm 0.092$ & \footnotesize$1.692 \pm 0.047$ & \footnotesize$1.688 \pm 0.046$ & 
 \footnotesize$1.683 \pm 0.044$\\

 \footnotesize $\epsilon = 0.4$ & \footnotesize$1.857 \pm 0.069$ & \footnotesize$1.721 \pm 0.090$ & \footnotesize$1.692 \pm 0.046$ & \footnotesize$1.690 \pm 0.045$ & 
 \footnotesize$1.684 \pm 0.0437$\\

 \footnotesize $\epsilon = 0.5$ & \footnotesize$1.859 \pm 0.068$ & 
 \footnotesize$1.721 \pm 0.090$ & \footnotesize$1.692 \pm 0.047$ & \footnotesize$1.689 \pm 0.046$ & 
 \footnotesize$1.684 \pm 0.044$\\

 \footnotesize $\epsilon = 0.6$ & \footnotesize$1.855 \pm 0.068$ & \footnotesize$1.719 \pm 0.088$ & \footnotesize$1.691 \pm 0.047$ & \footnotesize$1.689 \pm 0.046$ & 
 \footnotesize$1.683 \pm 0.044$\\

 \footnotesize $\epsilon = 0.7$ & \footnotesize$1.854 \pm 0.068$ & \footnotesize$1.717 \pm 0.087$ & \footnotesize$1.691 \pm 0.047$ & \footnotesize$1.688 \pm 0.046$ & 
 \footnotesize$1.683 \pm 0.043$\\

 \footnotesize $\epsilon = 0.8$ & \footnotesize$1.718 \pm 0.087$ & \footnotesize$1.718 \pm 0.076$ & \footnotesize$1.693 \pm 0.047$ & \footnotesize$1.689 \pm 0.045$ & 
 \footnotesize$1.683 \pm 0.043$\\

 \footnotesize $\epsilon = 0.9$ & \footnotesize$1.853 \pm 0.068$ & \footnotesize$1.716 \pm 0.086$ & 
 \footnotesize$1.692 \pm 0.046$ & \footnotesize$1.689 \pm 0.048$ & 
 \footnotesize$1.684 \pm 0.043$\\

 \footnotesize $\epsilon = 1.0$ & \footnotesize$1.852 \pm 0.069$ & \footnotesize$1.718 \pm 0.086$ & \footnotesize$1.691 \pm 0.047$ & \footnotesize$1.687 \pm 0.045$ & 
 \footnotesize$1.685 \pm 0.043$\\

\addlinespace
\bottomrule
\end{tabular*}}
\caption{Ablation study of the PEHE Errors for the IHDP dataset. The values in bold denote the least error across values for $\epsilon$ for a specific value of  $\alpha$.}
\label{tab:ablationihdptablepehe}
\end{table*}
\endgroup
\section{Hyperparameters}
We describe the various hyperparameters used in all our experiments here for all the methods which have been used for the comparison of results. Firstly, we look at the tabular baseline methods, i.e., T-learner, S-learner, X-Learner, DA-Learner (Domain Adaptation), and DR-Learner (Doubly Robust) \cite{econml, kunzel2019metalearners}:
\begin{itemize}
    \item Firstly, across all datasets, after inducing confounding, we do a 80/20 train-test split.
    \item Across all experiments, the number of estimators, i.e., `\textit{n\_estimators}' is set to $100$.
    \item The `\textit{max\_depth}' is set to $6$ and the `\textit{min\_samples\_leaf}' is set to $1$.
    \item $k$ is set to 5 for cross validation for the DR-Learner.
\end{itemize}

For neural architectures such as CFRNet, TARNet, FISH as well as our methods, we use pre-determined hyperparameters \cite{shalit2017estimating, Asami} -- which we describe first as follows:
\begin{itemize}
    \item Number of epochs for the \textit{ihdp} and \textit{jobs} datsets is $1000$ for both the representation layer and the CFR architecture and $250$ and $25$ resepctively for the representation layer and the CFR architecture for the \textit{cattaneo} dataset.
    \item \textit{weight\_decay} for the optimizer is 0.5 and \textit{gamma} for the scheduler is set to 0.97
    \textit{sigma} for Radial Basis Function (RBF) Maximum Mean Discrepancy (MMD) denoted as \textit{mmd\_rbf} for the IPM term is set to 0.1. However, we use Linear MMD (\textit{lin\_mmd}) in our experiments. 
    \item Learning rates for the \textit{ihdp}, \textit{jobs}, and the \textit{cattaneo} datasets is $10^{-4}$.
    \item The representation network, \textit{repnet} and the network responsible for splitting the hypothesis, \textit{outnet} both have $3$ layers, a hidden layer dimension of $48$ and a dropout rate of $0.145$
    \item For CFR-based baselines, we use $\alpha = 100000$ and for TAR-based baselines, $\alpha = 0$.
    \item For FISH, we fixed $\epsilon = 0.5$.
\end{itemize}

We perform experiments to fix our hyperparameters appropriately. We perform an ablation study to fix appropriate values for $\alpha$ and $\epsilon$:
\begin{itemize}
    \item For \textit{ihdp} and \textit{jobs} datsets, the combination for CFR is $\alpha = 10$ and $\epsilon = 0.9$ and TAR is $\alpha = 0$ (fixed) and $\epsilon = 0.8$. For the \textit{cattaneo} dataset, the combination for CFR is $\alpha = 10^{6}$ and $\epsilon = 0.8$ and TAR is $\alpha = 0$ (fixed) and $\epsilon = 0.8$.
\end{itemize}

Hyperparameters which we used for the image dataset \cite{takeuchi2021grab} were set as follows:
\begin{itemize}
    \item The hyperparameter for the outcome variable -- \textit{evacuation time} was set to `\textit{MAX}'. Hence, the experiments and simulations were run to reduce the maximum evacuation time taken. 
    \item \textit{hidden\_rep} was set to the default value 10 in the convolution layer.
    \item \textit{c\_rep} was set to \{10, 10\} and \textit{c\_out} was set to \{80, 50\}.
    \item $\alpha$ was set to $10^{6}$ for CFRConv and $0$ for TARConv.
    \item The \textit{learning rate} and the \textit{weight\_decay} were set to $10^{-2}$ and $10^{-8}$ respectively.
\end{itemize}
For further details on the implementation, kindly refer to the code files provided in the supplementary material.

\end{document}